\documentclass[10pt,twocolumn,letterpaper]{article}

\usepackage{iccv}
\usepackage{times}
\usepackage{epsfig}
\usepackage{graphicx}
\usepackage{amsmath}
\usepackage{amssymb}
\usepackage{multirow}
\usepackage{bm}
\newtheorem{Proposition}{Proposition}

\usepackage[breaklinks=true,bookmarks=false]{hyperref}

\iccvfinalcopy 

\def\iccvPaperID{****} 

\ificcvfinal\fi
\begin{document}

\title{Correntropy Induced L2 Graph for Robust Subspace Clustering}

\author{Canyi Lu$^1$, Jinhui Tang$^2$, Min Lin$^1$, Liang Lin$^3$, Shuicheng Yan$^1$, Zhouchen Lin$^{4,}$\thanks{Corresponding author.}\\
$^1$ Department of Electrical and Computer Engineering, National University of Singapore\\
$^2$ School of Computer Science, Nanjing University of Science and Technology\\
$^3$ School of Software, Sun Yat-Sen University\\
$^4$ Key Laboratory of Machine Perception (MOE), School of EECS, Peking University\\
{\tt\small canyilu@gmail.com, jinhuitang@mail.njust.edu.cn, mavenlin@gmail.com}\\
{\tt\small linliang@ieee.org, eleyans@nus.edu.sg, zlin@pku.edu.cn}
}

\maketitle
\thispagestyle{empty}

\begin{abstract}
In this paper, we study the robust subspace clustering problem, which aims to cluster the given possibly noisy data points into their underlying subspaces. A large pool of previous subspace clustering methods focus on the graph construction by different regularization of the representation coefficient. We instead focus on the robustness of the model to non-Gaussian noises. We propose a new robust clustering method by using the correntropy induced metric, which is robust for handling the non-Gaussian and impulsive noises. Also we further extend the method for handling the data with outlier rows/features. The multiplicative form of half-quadratic optimization is used to optimize the non-convex correntropy objective function of the proposed models. Extensive experiments on face datasets well demonstrate that the proposed methods are more robust to corruptions and occlusions.
\end{abstract}

\section{Introduction}
\label{sec:intro}  

\begin{figure}[!t]
\centering
\includegraphics[width=0.35\textwidth]{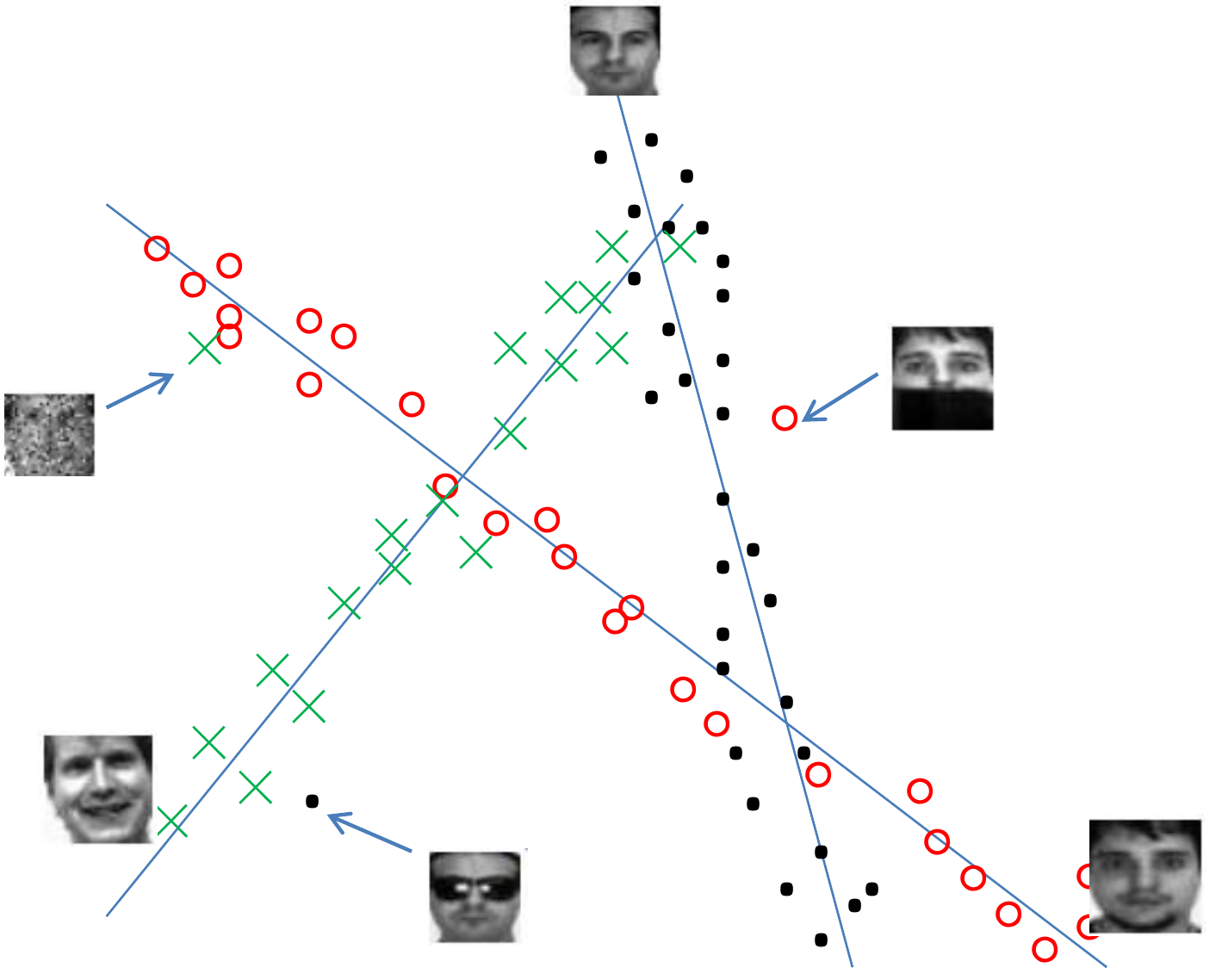}
\caption{Face images belonging to different subjects lie in different subspaces. Noises by and corruptions deviate the data from the underlying subspaces.}
\label{fig_subspaces}
\end{figure}

In pattern recognition and computer vision community, the data usually follow certain type of simple structure that enables intelligent representation. The subspaces are possibly the most widely used data model, since many real-world data, such as face images and motions, can be well characterized by subspaces. Given a set of data points, assuming that they are drawn from multiple subspaces, the goal of subspace clustering is to (1) cluster these data points into clusters with each cluster corresponding to a subspace, and (2) predict the memberships of the subspaces, including the number of subspaces  and the basis of each subspace. Subspace clustering is a fundamental problem and has numerous applications in the machine learning and computer vision literature, e.g. motion segmentation \cite{tron2007benchmark} and image clustering \cite{LSR}. The challenge in these applications lies in that the only known information is the data points, and they are usually contaminated by various noises. Figure \ref{fig_subspaces} illustrates some face images from three subjects. The face images with pixel corruption, sunglasses and/or scarf, deviate from their underlying subspaces. In this case, the subspace clustering is challenging. This paper aims to address the robust subspace clustering problem with various noises, such as the non-Gaussian noises.

\subsection{Summary of Main Notations}
In this work, matrices are represented with capital symbols. In particular, $I$ denotes the identity matrix. For a matrix $M$, $M_{ij}$ and $(M)_{ij}$ denote its $(i,j)$-th entry. $M^i$ is its $i$-th row, and $M_j$ is its $j$-th column. $\text{Diag}(v)$ converts the vector $v$ into a diagonal matrix in which the $i$-th diagonal entry is $v_i$. $\mathbb{R}_+$ denotes the set of non-negative real values and $\mathbb{S}_+^{d\times n}$ denote the set of positive semi-definite matrices. $M\succ0$ denotes that $M$ is symmetric and positive definite. $C^1$ denotes the set of continuous first derivative functions.

$||v||_2$ and $||v||_\infty$ denote the L2 norm and infinity norm of vector $v$, respectively. L1 norm, L21 norm and nuclear norm of matrix $M$ are defined as $||M||_1=\sum_{ij}|M_{ij}|$, $||M||_{21}=\sum_{j}||M_{j}||_2$, and $||M||_*=\sum_i\sigma_i$ ($\sigma_i$ is the $i$-th singular value of $M$), respectively.

\subsection{Related Work}
Many subspace clustering methods have been proposed \cite{tron2007benchmark,LRR,fang2012graph,gui2012discriminant}. In this work, we focus on the recent graph based subspace clustering methods \cite{L1Graph,SSC,LRR,LRRpami,LSR}. These methods are based on the spectral clustering, and its first step aims to construct an affinity (or graph) matrix which is close to be block diagonal, with zero elements corresponding to data pair from different subspaces. After the affinity matrix is learned, the Normalized Cut \cite{NormaCut} is employed to segment the data into multiple clusters. For a given data matrix $X\in\mathbb{R}^{d\times n}$, where $d$ denotes the feature dimension and $n$ is the number of data points, the most recent methods, including L1-graph \cite{L1Graph} or Sparse Subspace Clustering (SSC) \cite{SSC}, Low-Rank Representation (LRR) \cite{LRR,LRRpami}, Multi-Subspace Representation (MSR) \cite{MSR} and Least Squares Representation (LSR) \cite{LSR} learn the affinity matrix $Z\in\mathbb{R}^{n\times n}$ by solving the following common problem
\begin{equation}
\label{Eq_gnd1}
\min_Z \mathcal{L}(X-XZ)+\lambda\mathcal{R}(Z).
\end{equation}

For L1-graph or SSC, $\mathcal{L}(X-XZ)=||X-XZ||_F^2$ and $\mathcal{R}(Z)=||Z||_1$. The motivation of using SSC is that the L1-minimization will lead to a sparse solution tending to be block diagonal. As pointed out in \cite{LSR}, the L1-minimization does not exhibit the grouping effect, and thus is weak to group correlated data points together.

For LRR, $\mathcal{L}(X-XZ)=||X-XZ||_{21}$ and $\mathcal{R}(Z)=||Z||_*$. It aims to find a low rank affinity matrix. When the data are drawn from independent subspaces, LRR leads to a bock diagonal solution which can recover the true subspaces. For the noisy case, LRR uses the robust L21-norm to remove outlier samples.

MSR simply combines the criteria of SSC and LRR, $\mathcal{L}(X-XZ)=||X-XZ||_{21}$ and $\mathcal{R}(Z)=||Z||_1+\gamma||Z||_*$. Thus MSR can be regarded as a tradeoff between SSC and LRR, but it needs to tune one more parameter $\gamma$.

The LSR method uses the Frobenius norm to model both the reconstruction error and the representation matrix,  $\mathcal{L}(X-XZ)=||X-XZ||_F^2$ and $\mathcal{R}(Z)=||Z||_F^2$. LSR has a closed form solution which makes it efficient, and the grouping effect makes it effective for subspace clustering.

The above methods share the common formulation as shown in (\ref{Eq_gnd1}). The Frobenius norm and L21 norm are used as the loss function while the L1 norm, nuclear norm and Frobenius are used to control the affinity matrix. Different formulations require different solvers for these problems. In this work, we show that the L1 norm, L21 norm and nuclear norm all satisfy certain conditions, and thus the previous subspace clustering methods, including SSC, LRR and MSR, can be unified within a general framework from the perspective of half-quadratic optimization \cite{HQanalysis}. The relationship between the general framework and the previous optimization methods for sparse and low rank minimization is also presented in this work.

Different from the previous methods which focus on a regularization term $\mathcal{R}(Z)$, this work focuses on the construction error term $\mathcal{L}(Z)$ for robust subspace learning. Previous works use the Frobenius norm to measure the quality of approximation, which is optimal for the case of independent and identically distributed (i.i.d.) Gaussian noise but not robust to outliers. LRR by using the L21 norm is able to remove the outlier samples, but it is sensitive to the outlier features. To overcome the weakness of mean squared error, we propose a new robust subspace clustering method which uses the correntropy induced metric as the loss function. The Frobenius norm is used to control the affinity matrix to preserve the grouping effect as in LSR. Then we minimize the non-convex correntropy objective of the proposed method by alternate minimization.

\subsection{Contributions and Organization}
We summarize the contributions of this work as follows:

\begin{itemize}
\item We propose a new robust subspace clustering method by Correntropy Induced L2 (CIL2) graph. It is able to handle data with non-Gaussian noises. We also extend CIL2 for handling data with outlier rows/features.
\item We apply the correntropy induced L2 graph for face clustering under various types of corruptions and occlusions. Extensive experiments demonstrate the effectiveness of the proposed method by comparing it with the state-of-the-art methods.
\end{itemize}

The remainder of this paper is organized as follows. Section 2 gives a brief review of the half-quadratic analysis and presents a general half-quadratic framework for robust subspace clustering. Section 3 elaborates the proposed CIL2 graph for robust subspace clustering. Section 4 provides experimental results on face clustering under different settings. We conclude this paper in Section 5.

\begin{table*}[!t]
\footnotesize
\caption{The popular previous subspace clustering models can be solved by half-quadratic minimization.}
\label{Tab_framework}
\centering
\begin{tabular}{|c |c |c c|}
\hline
\multirow{2}*{ Methods } & { Objective }  & \multicolumn{2}{c|}{Function}  \\
 & $\min_{Z}\mathcal{L}(X-XZ)+\lambda\mathcal{R}(Z)$ &  $\mathcal{L}(\cdot)$ & $\mathcal{R}(\cdot)$  \\ \hline
SSC \cite{SSC} & $\min_Z ||X-XZ||_F^2+\lambda||Z||_1$ & $||\cdot||_F^2$ & $||\cdot||_1$  \\ \hline
LRR \cite{LRR} & $\min_Z ||X-XZ||_{21}+\lambda||Z||_*$ & $||\cdot||_{21}$ & $||\cdot||_*$  \\ \hline
MSR \cite{MSR}  & $\min_Z ||X-XZ||_{21}+\lambda||Z||_1+\lambda\gamma||Z||_*$ & $||\cdot||_{21}$ & $||\cdot||_1+\gamma||\cdot||_*$ \\ \hline
LSR \cite{LSR} & $\min_Z ||X-XZ||_F^2+\lambda||Z||_F^2$ & $||\cdot||_F^2$ & $||\cdot||_F^2$  \\ \hline
\end{tabular}
\end{table*}

\section{A General Half-Quadratic Framework for Robust Subspace Clustering}

For a given data matrix $X\in\mathbb{R}^{d\times n}$, consider the following general problem:
\begin{equation}
\label{Eq_gnd}
\begin{split}
\min_{Z} \ & \mathcal{J}(Z)=\mathcal{L}(E)+\lambda\mathcal{R}(Z)\\
\text{s.t. } \ & E=X-XZ,
\end{split}
\end{equation}
where $\mathcal{L}(E)$ is the loss function chosen to be robust to outliers or gross errors, and $\mathcal{R}(Z)$ is the regularization term. The loss function $\mathcal{L}(E)$ and regularization $\mathcal{R}(Z)$ may be non-quadratic. Thus it may be difficult to solve the problem (\ref{Eq_gnd}). But if $\mathcal{L}(E)$ and $\mathcal{R}(Z)$ satisfy certain conditions, we can minimize $\mathcal{J}(Z)$ by half-quadratic analysis.

In this work, we consider a general case of $\phi(x)$ that satisfies the following conditions \cite{HQanalysis}
\begin{equation}
\label{Eq_cond}
\begin{split}
&\text{(a) } x\rightarrow \phi(x) \text{ is convex on} \ \mathbb{R},\\
&\text{(b) } x\rightarrow \phi(\sqrt{x}) \text{ is concave on} \ \mathbb{R}_+, \\
&\text{(c) } \phi(x)=\phi(-x), x\in\mathbb{R}, \\
&\text{(d) } \phi(x) \text{ is } C^1 \text{ on } \mathbb{R}, \\
&\text{(e) } \phi''(0^+)>0,\\
&\text{(f) } \lim_{x\rightarrow \infty}\phi(x)/x^2=0.
\end{split}
\end{equation}
Or in the matrix form $\phi(M)$:
\begin{equation}
\label{Eq_condmatrix}
\begin{split}
&\text{(a) } M\rightarrow \phi(M) \text{ is convex on} \ \mathbb{R}^{N\times N},\\
&\text{(b) } M\rightarrow \phi(\sqrt{M}) \text{ is concave on} \ \mathbb{S}_+^{N\times N}, \\
&\text{(c) } \phi(M)=\phi(-M), \ M\in\mathbb{R}^{N\times N}, \\
&\text{(d) } \phi(M) \text{ is } C^1 \text{ on } \mathbb{R}^{N\times N}, \\
&\text{(e) } \phi(M) \text{ is strictly convex on } 0,\\
&\text{(f) } \lim_{M\rightarrow \infty}\phi(M)/||M||_F^2=0.
\end{split}
\end{equation}

If $\phi(\cdot)$ satisfies all the conditions in (\ref{Eq_cond}), there exists a dual function $\psi$  \cite{HQanalysis} such that
\begin{equation}
\label{Eq_dual}
\phi(x)=\inf_{s\in\mathbb{R}}\{\frac{1}{2}sx^2+\psi(s)\},
\end{equation}
where $s$ is determined by the minimizer function $\delta(\cdot)$ with respect to $\phi(\cdot)$. $\delta(\cdot)$ admits an explicit form under certain restrictive assumptions:
\begin{equation}
\label{Eq_miner}
s=\delta(t)=
\begin{cases}
\phi''(0^+), \quad &\text{if }t=0, \\
\frac{\phi'(t)}{t}, \quad &\text{if } t\neq 0.
\end{cases}
\end{equation}

If $\mathcal{L}(E)=\sum_{ij}\phi(E_{ij})$ (similar analysis can be performed on $\mathcal{R}(Z)$), problem (\ref{Eq_gnd}) reads:
\begin{equation}
\label{Eq_genhq11}
\begin{split}
\min_Z\mathcal{J}(Z) &=  \sum_{ij}\phi(E_{ij})+\lambda\mathcal{R}(Z)\\
\text{s.t.} &\ E=X-XZ.
\end{split}
\end{equation}
Using (\ref{Eq_genhq11}) on each $E_{ij}$, the augmented function of $\mathcal{J}$ of (\ref{Eq_genhq11}) is as follows
\begin{equation}
\label{Eq_genhq}
\begin{split}
\mathcal{J}(Z,S) = \sum_{ij}(\frac{1}{2}S_{ij}E_{ij}^2+\psi(S_{ij}))+\lambda\mathcal{R}(Z).
\end{split}
\end{equation}

Based on the half-quadratic optimization, $\mathcal{J}(Z,S)$ can be minimized by the following alternate procedure:
\begin{equation}
\label{Eq_hq1}
S_{ij}=\delta(E_{ij}),
\end{equation}
\begin{equation}
\label{Eq_hq2}
Z=\arg\min_{Z}\sum_{{ij}}\frac{1}{2}S_{ij}E_{ij}^2+\lambda\mathcal{R}(Z).
\end{equation}
The update sequence generated by the above scheme will converges. The objective function in (\ref{Eq_genhq}) is nonincreasing under the update rules in (\ref{Eq_hq1})(\ref{Eq_hq2}) \cite{HQanalysis}.

For L1 norm, $\phi_1(x)=|x|=\sqrt{x^2}$ does not satisfy condition (d) in (\ref{Eq_cond}). We use $\phi_1(x)=\sqrt{x^2+\epsilon^2}$ as an approximation of $|x|$ with a small positive value $\epsilon$. It can be easily seen that $\sqrt{x^2+\epsilon^2}$ satisfies all the conditions in (\ref{Eq_cond}). We roughly say the L1 norm satisfies all the conditions in (\ref{Eq_cond}) in this sense. Previous work \cite{chartrand2008iteratively} for solving the L1-minimization by iteratively reweighted least squares optimization can be interpreted as the half-quadratic optimization in (\ref{Eq_hq1}) and (\ref{Eq_hq2}).
For L21 norm, $\phi_{21}(X)=||X||_{21}=\sum_i||X_i||_2\approx\sum_i(||X_i||_2^2+\epsilon)^{\frac{1}{2}}$, where $\epsilon$ is a small positive value. It is easy to check that $\phi_{21}(x)=(x^2+\epsilon)^{\frac{1}{2}}$ also satisfies all the conditions in (\ref{Eq_cond}).
For nuclear norm, $\phi_*(X)=\text{Tr}(X^TX)^{\frac{1}{2}}\approx\text{Tr}(X^TX+\epsilon I)^{\frac{1}{2}}$, where $\epsilon$ is a small positive value. It is easy to check that $\text{Tr}(X^TX+\epsilon I)^{\frac{1}{2}}$ satisfies the conditions (a)-(e) in (\ref{Eq_condmatrix}). For the condition (f), the $i$-th singular value $\sigma_i$ of $X$ converges to infinity when $X\rightarrow\infty$, and thus $\lim_{X\rightarrow\infty}\phi_*(X)/||X||_F^2=\lim_{\sigma_i\rightarrow\infty}\frac{\sum_i\sigma_i}{\sum_i\sigma_i^2}=0$. Therefore the nuclear norm also satisfies all the conditions in (\ref{Eq_condmatrix}). The work \cite{IRLSrank} for solving low rank minimization by iteratively reweighted least squares minimization can be interpreted as the half-quadratic minimization.

If both two functions satisfy all the  conditions in (\ref{Eq_cond}), the sum of them also satisfies these conditions. The optimization method in \cite{MSR} for minimizing  $||X||_1+\gamma||X||_*$ can be regarded as the half-quadratic optimization in (\ref{Eq_hq1})(\ref{Eq_hq2}).

Based on the above analysis, previous subspace clustering methods by using the L1 norm, L21 norm and nuclear norm can be optimized by the half-quadratic analysis on (\ref{Eq_hq1})(\ref{Eq_hq2}) by slightly relaxing the objective function. As shown in Table \ref{Tab_framework}, previous subspace clustering methods, including SSC, LRR, MSR and LSR, can be regarded as special cases of the problem (\ref{Eq_gnd}) from the view of half-quadratic analysis. Note that the Frobenius norm $||\cdot||_F^2$ does not need half-quadratic representation because it is already quadratic. We also list it in Table \ref{Tab_framework} since it is widely used.

\begin{figure}[!t]
\centering
\includegraphics[width=0.35\textwidth]{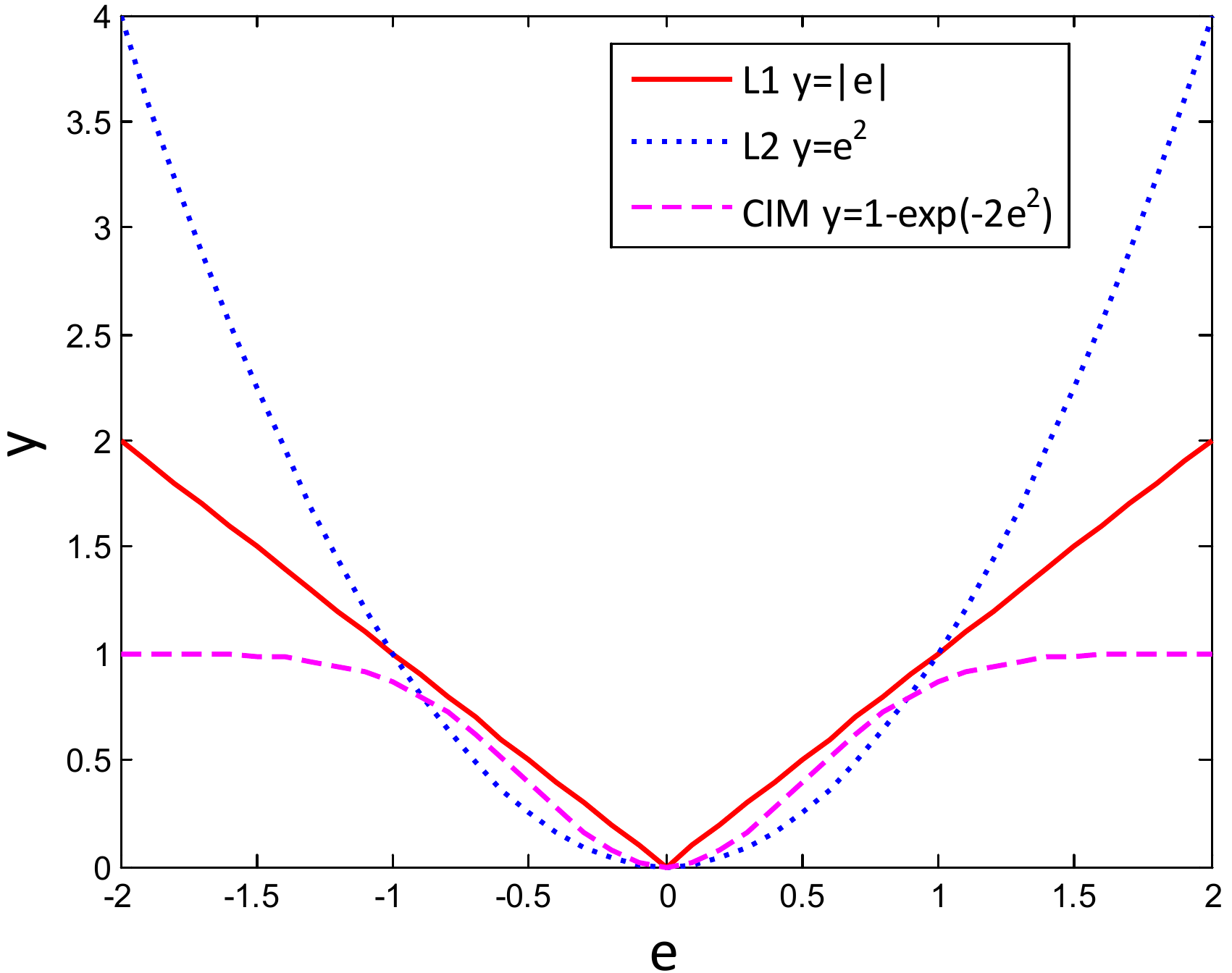}
\caption{Comparison of different loss functions.}
\label{fig_l1l2cim}
\end{figure}

\section{Correntropy Induced L2 Graph for Robust Subspace Clustering}

\subsection{Correntropy Induced Metric}

The mean squared errors (MSE) are probably the most widely used methodologies for quantifying how similar two random variables are. Successful engineering solutions from this methodology rely heavily on the Gaussianity and linearity assumptions. The work in \cite{erdogmus2002error} extended the concept of mean squared error adaptation to information theoretic learning (ITL) to include the information theoretic criteria. Then they further proposed the concept of correntropy to process non-Gaussian and impulsive noises \cite{liu2007correntropy}. The correntropy is a generalized similarity measure between two arbitrary scalar random variables $u$ and $v$ defined by
\begin{equation}
V_{\sigma}(u,v)=\textbf{E}[k_\sigma(e)],
\end{equation}
where $e=u-v$, $\textbf{E}[\cdot]$ is the expectation operator, and $k_\sigma(\cdot)$ is the kernel function. In this work we only consider the Gaussian kernel $k_\sigma(e)=\exp(-e^2/2\sigma^2)$. In practice, we usually have only a finite number
of data $\{(u_i,v_i)\}_{i=1}^n$, which leads to the sample estimator of correntropy:
\begin{equation}
\label{Eq_corr}
\hat{V}_{\sigma}(u,v)=\frac{1}{n}\sum_{i=1}^nk_{\sigma}(u_i-v_i).
\end{equation}

Based on (\ref{Eq_corr}), Liu et al. \cite{liu2007correntropy} extended the concept of correntropy criterion for a general similarity measurement between any two vectors, which is called the Correntropy Induced Metric (CIM). It is formally defined as
\begin{equation}
\text{CIM}(u,v)=(k(0)-\frac{1}{n}\sum_{i=1}^nk_\sigma(e_i))^{1/2},
\end{equation}
where $e_i=u_i-v_i$, for each $i=1,\cdots,n$.

Figure \ref{fig_l1l2cim} shows a comparison of the absolute error, mean squared error and CIM. The mean squared error is a global metric which increases quadratically for large errors. CIM is a local metric which is close to the absolute error when the errors are relatively small. For large errors, the value of CIM is close to 1. Note that the large errors are usually caused by outliers, but their effect on CIM is limited. Therefore CIM will be more robust to the non-Gaussian noises. The effectiveness and robustness of correntropy have been verified in face recognition \cite{MMC}, feature selection \cite{L21RFS} and signal processing \cite{liu2007correntropy}. This paper uses this concept for robust subspace clustering.

\subsection{Correntropy Induced L2 Graph}

For robust subspace clustering, we use the correntropy to replace the Frobenius norm in the LSR model to model the reconstruction error, leading to the Correntropy Induced L2 (CIL2) graph as follows:
\begin{equation}
\label{Eq_CIL2}
\begin{split}
\min_Z &\sum_{i,j}(1-k_\sigma(E_{ij}))+\lambda||Z||_F^2\\
\text{s.t.} \ & \  E=X-XZ.
\end{split}
\end{equation}
It is easy to check that $\phi_\sigma(x)=1-k_\sigma(x)=1-\exp(-x^2/2\sigma^2)$ satisfies all the conditions in (\ref{Eq_cond}). Therefore the above problem can be solved by the half-quadratic analysis. According to (\ref{Eq_genhq}), problem (\ref{Eq_CIL2}) is equivalent to the following augmented objective function:
\begin{equation}
\label{Eq_cil2J}
\begin{split}
\mathcal{J}(Z,S) = & \sum_{ij}(\frac{1}{2}S_{ij}E_{ij}^2+\psi(S_{ij}))+\lambda||Z||_F^2\\
\text{s.t.} \ & \  E=X-XZ,
\end{split}
\end{equation}
where $\psi(\cdot)$ is the dual function corresponding to $\phi_\sigma(\cdot)$. We can minimize $\mathcal{J}(Z,S)$ in (\ref{Eq_cil2J}) by the following alternate procedure:

\begin{equation}
\label{Eq_cil2hq1}
S_{ij}=\frac{1}{\sigma^2}\exp(-E_{ij}^2/2\sigma^2),
\end{equation}
\begin{equation}
\label{Eq_cil2hq2}
Z_i=\arg\min_{Z_i}(X-XZ)_i^T\text{Diag}(S_i)(X-XZ)_i+\lambda||Z_i||_2^2.
\end{equation}
Let $\hat{X}=\text{Diag}(\sqrt{S_i})X$, then problem (\ref{Eq_cil2hq2}) is also a least square regression model:
\begin{equation}
\label{Eq_cil2lsr}
\min_{Z_i}||\hat{X}-\hat{X}Z_i||_2^2+\lambda||Z_i||_2^2.
\end{equation}

Since the kernel size $\sigma$ may affect the performance of the proposed model. It is usually determined empirically. In this study, the kernel size is computed as the average reconstruction error,
\begin{equation}
\sigma^2=\frac{1}{2dn}||X-XZ||_F^2.
\end{equation}

From (\ref{Eq_cil2J}) or problem (\ref{Eq_cil2lsr}), we can see that the correntropy based LSR model can be regarded as a weighted LSR, where each weight $S_{ij}$ corresponding to $E_{ij}$ is used to control the effect of $E_{ij}$.


\subsection{Row Based Correntropy Induced L2 Graph}

In some real-world applications, the data may be occluded with outlier rows/features. For example, some rows of the face images with sunglasses and scarf are outliers, which are not discriminative for classification and clustering. In this case, we should measure the quality of the reconstruction error based on the entire row. The effect of rows can be controlled by assigning different weights, and each element in the same row has the same weight. To this end, we have the row based Correntropy Induced L2 (rCIL2) graph by solving the following problem
\begin{equation}
\label{Eq_rCIL2}
\begin{split}
\min_Z &\sum_{i}(1-k_\sigma(||E^i||_2))+\lambda||Z||_F^2\\
\text{s.t.} \ & \  E=X-XZ.
\end{split}
\end{equation}
According to the half-quadratic analysis, the above problem is equivalent to the following problem
\begin{equation}
\label{Eq_rCIL2hq}
\mathcal{J}_r(Z,w)=\sum_i(\frac{1}{2}w_i||X^i-X^iZ||_2^2+\psi(w_i))+\lambda||Z||_F^2.
\end{equation}
Problem (\ref{Eq_rCIL2hq}) can be solved by updating $Z$, $w$, and $\sigma$ alternately as follows:
\begin{equation}
w_i=\frac{1}{\sigma^2}\exp(-(X^i-X^iZ)^2/2\sigma^2),
\end{equation}
\begin{equation}
\label{Eq_ger}
Z=\arg\min_Z \text{Tr}((X-XZ)^T\text{Diag}(w)(X-XZ))+\lambda||Z||_F^2,
\end{equation}
\begin{equation}
\sigma^2=\frac{1}{2d}\sum_i||X^i-X^iZ||_2^2.
\end{equation}

According to (\ref{Eq_miner}) and (\ref{Eq_hq2}), it is easy to prove that the sequences $\{\hat{\mathcal{J}}(Z^t,S^t),t=1,2,\cdots\}$ in (\ref{Eq_cil2J}) and $\{\hat{\mathcal{J}}(Z^t,w^t),t=1,2,\cdots\}$ in (\ref{Eq_rCIL2hq}) converge.

\subsection{The Grouping Effect}
The CIL2 and rCIL2 graphs also use the L2 regularization as in LSR \cite{LSR}. It is expected that they also have the grouping effect, i.e. the coefficients of a group of correlated data are approximately equal. The obtained solutions by CIL2 in (\ref{Eq_cil2hq2}) and by rCIL2 in (\ref{Eq_ger}) are the weighted least square regression model which owns the grouping effect:

\begin{Proposition}
\label{thm_groupingeffect}
Given a data vector $y\in\mathbb{R}^d$, data points $X\in\mathbb{R}^{d\times n}$, the weight vector $w\in\mathbb{R}^d$ corresponding to each row of $X$, and a parameter $\lambda$. Assume that each data point of $X$ is normalized. Let $z^*$ be the optimal solution to the following weighted LSR (in vector form) problem:
\begin{equation}
\label{Eq_LSRvector}
\min_z ||\text{Diag}(w)(y-Xz)||_2^2+\lambda ||z||_2^2.
\end{equation}
We have
\begin{equation}
\label{Eq_groupingeffect}
\frac{||z_i^*-z_j^*||_2}{||w||_2||\text{Diag}(w)y||_2}\leq\frac{1}{\lambda}\sqrt{2(1-r)},
\end{equation}
where $r=X_i^TX_j$ is the sample correlation.
\end{Proposition}

We omit the proof of Proposition \ref{thm_groupingeffect} here, it can be proved in the same way as the Theorem 7 in \cite{LSR}.

The mechanism of correntropy and the Proposition \ref{thm_groupingeffect} ensure that both CIL2 and rCIL2 are not only robust to noises but also preserve the grouping effect.

\subsection{Algorithm for Subspace Clustering}
Similar to the previous subspace clustering method LSR, which uses the representation coefficient matrix to construct the graph for clustering, we apply the learned solution $Z^*$ by CIL2 and rCIL2 to construct a graph with weights $W=(|Z^*|+|{Z^*}^T|)/2$, and then Normalized Cut \cite{NormaCut} is applied to cluster the data points into multiple clusters.

\begin{figure}[!t]
\centering
\includegraphics[width=0.45\textwidth]{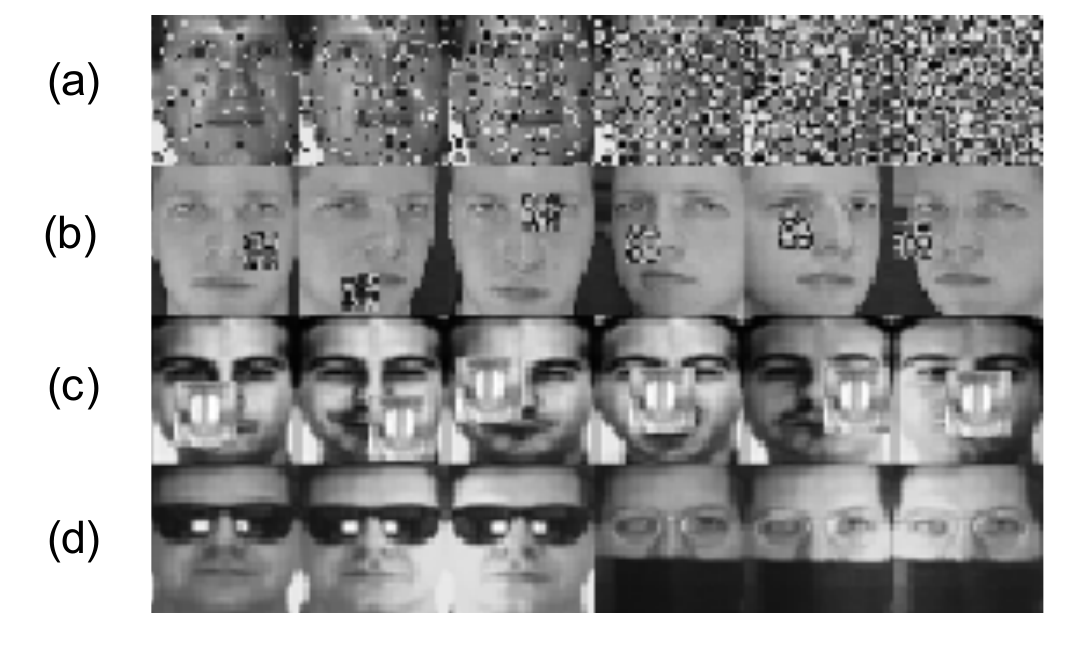}
\caption{ (a) Some corrupted face images from the Yale dataset, with 10$\%$, 20$\%$ 30$\%$, 50$\%$, 70$\%$ and 90$\%$ of pixels corrupted, respectively; (b) Some face images with random block occlusion from the ORL dataset; (c) Some face images with 20$\%$ occlusion by monkey face from the AR dataset; (d) Some face images with contiguous occlusion by sunglasses and scarf from the AR dataset.}
\label{fig_faceimages}
\end{figure}

\section{Experiments}

\subsection{Datasets and Settings}

Our experiments are performed on three face datasets: Yale, ORL and AR. Descriptions of these data sets are given as follows.

The Yale face dataset \cite{belhumeur1997eigenfaces} contains 165 grayscale images of 15 individuals.  The images demonstrate variations in lighting condition and facial expression (normal, happy, sad, sleepy,  surprised, and wink). The grayscale images are resized to a resolution of $32\times 32$ pixels.

The ORL face dataset \cite{samaria1994parameterisation} contains 400 images of 40 individuals. Some images were captured at different times and have different variations including expression (open or closed eyes, smiling or non-smiling) and facial details (glasses or no glasses). The images were taken with a tolerance for some tilting and rotation of the face up to 20 degrees. Each image is resized to $32\times 32$ pixels.

The AR database \cite{AR} consists of over 4,000 facial images from 126 subjects. For each subject, 26 facial images are taken in two separate sessions. These images suffer different facial variations, including various facial expressions (neutral, smile, anger, and scream), illumination variations (left light on, right light on, and all side lights on), and occlusion by sunglasses or scarf. We select a subset of the data set consisting of 50 male subjects and 50 female subjects. The grayscale images are resized to a resolution of $32\times 32$ pixels.
\begin{figure}[!t]
\centering
\includegraphics[width=0.45\textwidth]{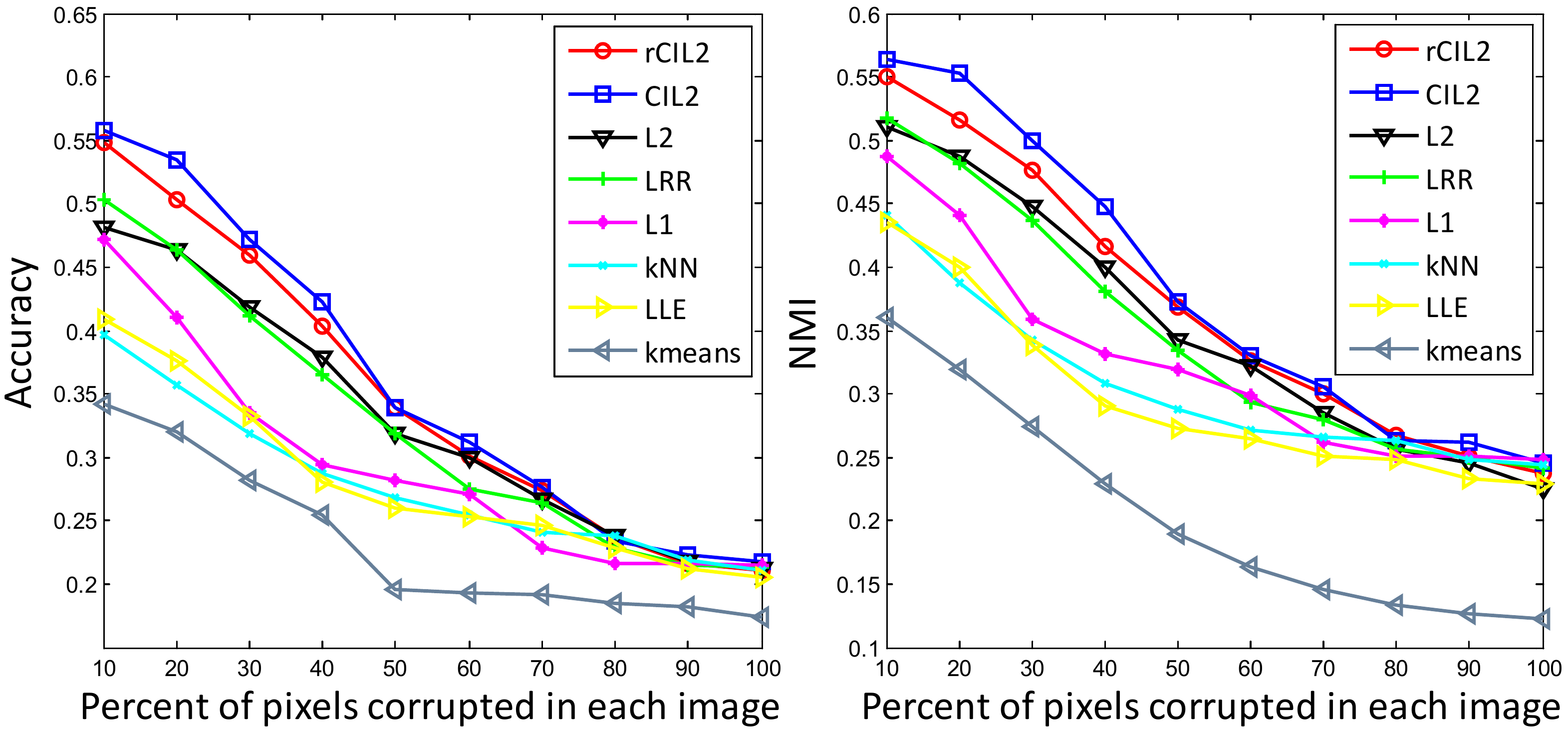}
\caption{Clustering accuracy and NMI on the Yale dataset with pixel corruption for different algorithms.}
\label{fig_yale_pixel}
\end{figure}

\subsection{Evaluation Metrics}

The clustering result is evaluated by the accuracy and normalized mutual information (NMI) metric as in\cite{xu2003document}. For each data point
$x_i$, let $p_i$ and $y_i$ be the obtained cluster label and the label
provided by the ground truth, respectively. The accuracy is defined as follows:
\begin{equation}
Accuracy=\frac{\sum_{i=1}^n\delta(y_i,map(p_i))}{n},
\end{equation}
where $\delta (a,b)$ is the delta function that equals one if $a=b$ and equals zero otherwise, and $map(p_i)$ is the permutation mapping function that maps each cluster label $p_i$ to the equivalent label in $y$.

Let $C$ denote the set of clusters obtained from the ground truth and $C'$ obtained by the segmentation method. Their mutual information metric $MI(C, C¡ä)$ is defined as follows:
\begin{equation}
MI(C,C')=\sum_{c_i\in C, c_j'\in C'}p(c_i,c_j')log_2\frac{p(c_i,c_j')}{p(c_i)p(cj')},
\end{equation}
where $p(c_i)$ and $p(c_j')$ are the probabilities that a sample point arbitrarily selected from the data point belongs to the clusters $c_i$ and $c_j'$, respectively, and $p(c_i, c_j')$ is the joint probability that the arbitrarily selected data point belongs to the clusters $c_i$ as well as $c_j'$ at the same time. We use the normalized mutual
information (NMI) as follows:
\begin{equation}
NMI(C,C')=\frac{MI(C,C')}{\max(H(C),H(C'))},
\end{equation}
where $H(C)$ and $H(C')$ are the entropies of $C$ and $C'$, respectively. It is easy to see that $NMI(C, C¡ä)$ ranges from 0 to 1. $NMI=1$ if the two sets of clusters are identical, and $NMI=0$ if the two sets are independent.

\subsection{Algorithm Settings}

We compare our rCIL2 and CIL2 graphs with several graph construction methods for subspace clustering, including the L1-graph \cite{L1Graph} (or SSC \cite{SSC}), L2-graph (LSR) \cite{LSR}, and LRR-graph \cite{LRRpami}. kNN and LLE \cite{LLE} are also applied to construct graphs for subspace clustering. Kmeans is used as the baseline for comparison. The model parameters of these methods are searched from the candidate value sets and the best results are reported.

\begin{figure}[!t]
\centering
\includegraphics[width=0.45\textwidth]{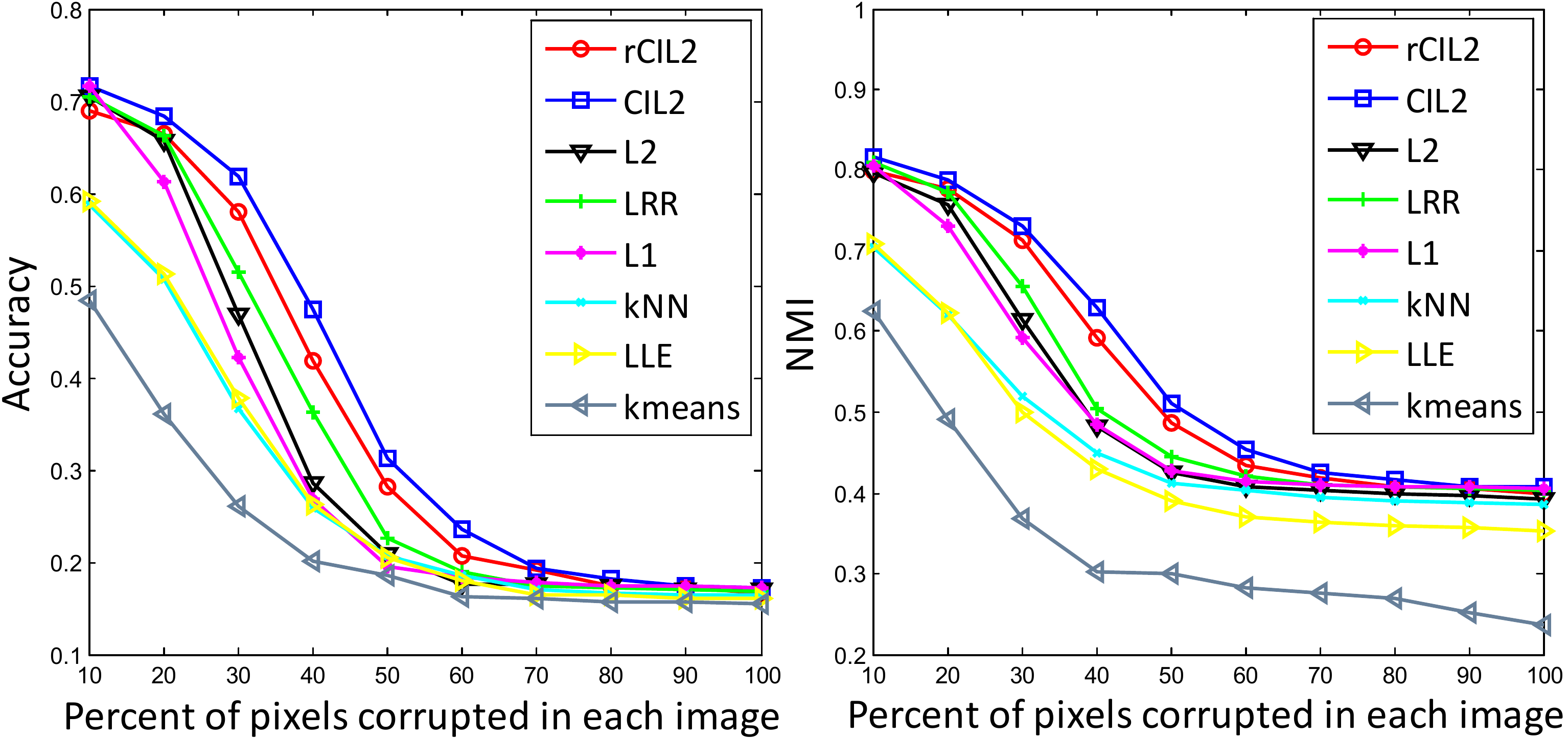}
\caption{Clustering accuracy and NMI on the ORL dataset with pixel corruption for different algorithms.}
\label{fig_orl_pixel}
\end{figure}

\subsection{Results under Random Pixel Corruption}

In some practical scenarios, the face images may be partially corrupted. We evaluate the algorithmic robustness on the Yale and ORL face datasets. Each image is corrupted by replacing a percentage of randomly chosen pixels with i.i.d. samples from a uniform distribution (uniform on [0, 255]). The corrupted pixels are randomly chosen for each image, and the locations are unknown. We vary the percentage $r$ of corrupted pixels from 10$\%$ to 100$\%$. Figure \ref{fig_faceimages} (a) shows some examples of those corruptions. To the human eyes, beyond 50$\%$ corruption, the corrupted images are barely recognizable as face images. Since the images are with random corruption, we repeat the experiments for 20 times for each $r$, and the means of accuracy and NMI are reported for evaluation.

Figures \ref{fig_yale_pixel} and \ref{fig_orl_pixel} show the means of clustering accuracy and NMI of different methods as functions of the corruption level. It can be found that both the accuracy and NMI decrease when more pixels of each image are randomly corrupted. Our proposed CIL2 and rCIL2 outperform the compared methods in most cases. In particular, the CIL2 usually performs better than rCIL2 when the percentage of the corrupted pixels is no more than 50$\%$ on the Yale dataset and 70$\%$ on the ORL dataset. This is because each row of images may not be regarded as outliers when the level of the random pixel corruption is low. LRR and L2-graph perform competitively on both datasets, which also verifies the effectiveness of the grouping effect of these two methods for subspace clustering. When the images are with high percentage of pixel corruptions, none of the compared methods perform well due to insufficient discriminative information.

\begin{figure}[!t]
\centering
\includegraphics[width=0.45\textwidth]{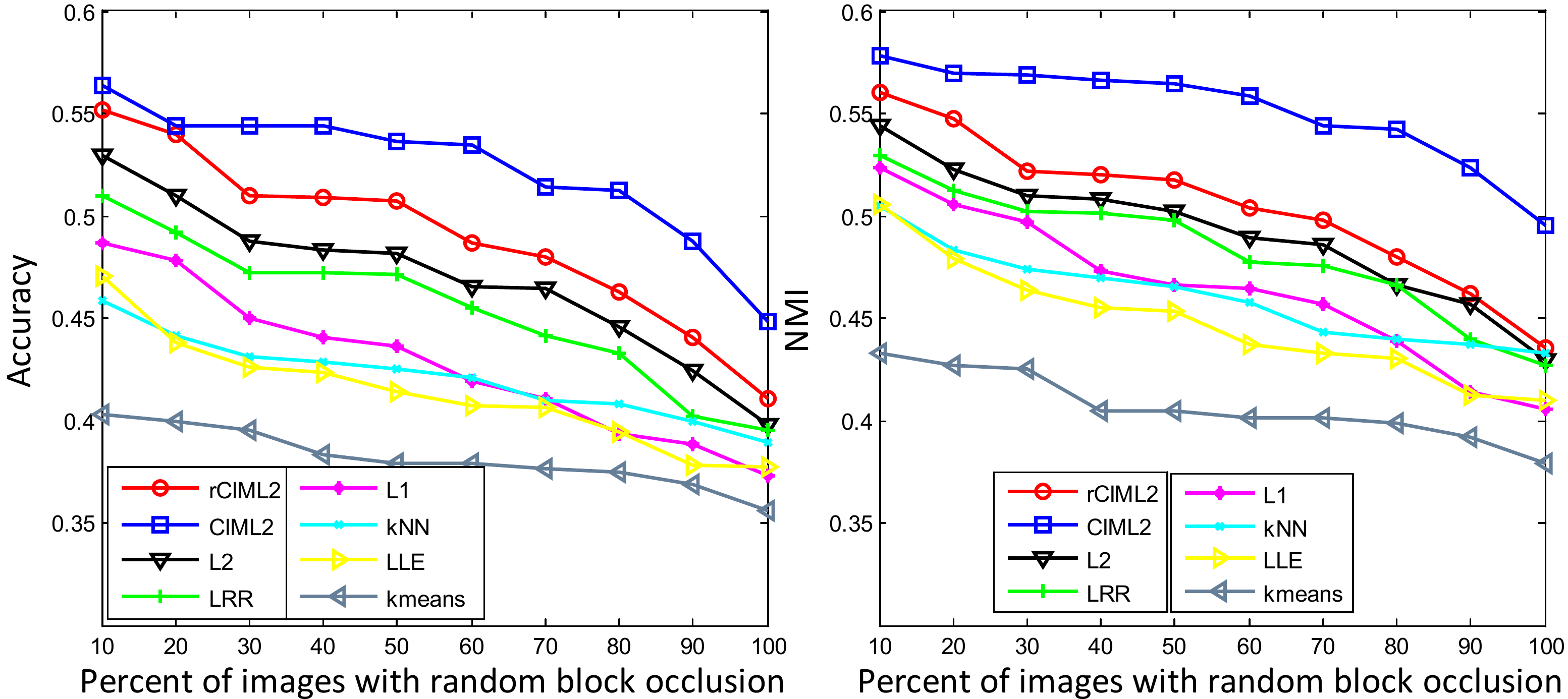}
\caption{Clustering accuracy and NMI on the Yale dataset with block occlusion for different algorithms.}
\label{fig_yale_block}
\end{figure}

\begin{figure}[!t]
\centering
\includegraphics[width=0.45\textwidth]{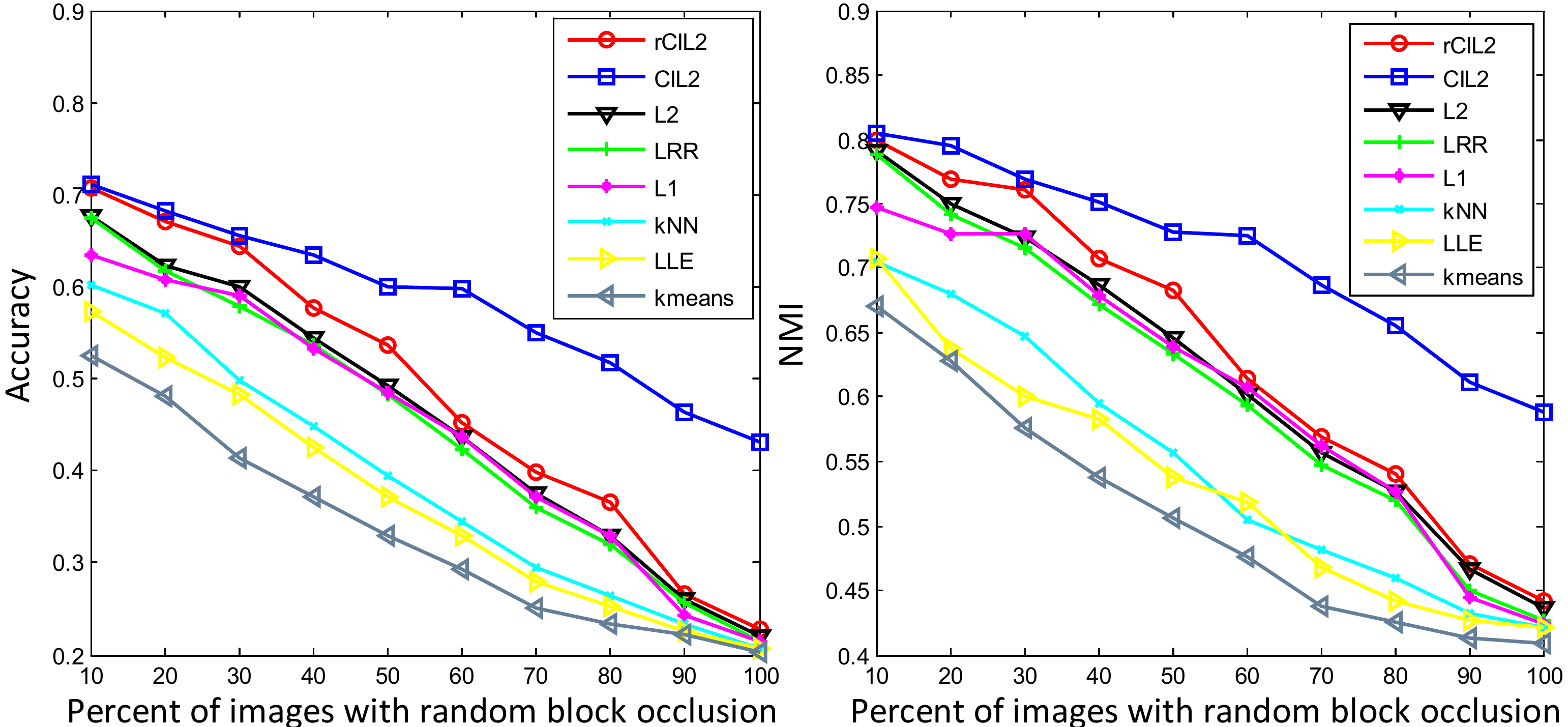}
\caption{Clustering accuracy and NMI on the ORL dataset with block occlusion for different algorithms.}
\label{fig_orl_block}
\end{figure}

\subsection{Results under Contiguous Occlusion}

In this subsection we simulate various types of contiguous occlusions by replacing a randomly selected local region in some randomly selected images with a black-white square and an unrelated monkey image.

The first experiment is conducted on the Yale and ORL datasets with random block occlusion. Figure \ref{fig_faceimages} (b) shows some face images with such black-white occlusions, in size of $8\times 8$ pixels. In each dataset, we select $r$ percentage of the images for occlusion, with $r$ varying from 10$\%$ to 100$\%$. The experiments are repeated 20 times for each $r$, and the means of accuracy and NMI are reported for evaluation.

Figures \ref{fig_yale_block} and \ref{fig_orl_block} show the means of clustering accuracy and NMI of each method on different percentages of corrupted images. CIL2-graph achieves the best accuracy and NMI on both Yale and ORL datasets in all cases. Compared with previous subspace clustering methods, the improvement by rCIL2 is still limited. The phenomenon is similar to the random pixel corruption scenario, since the images with block occlusion will not lead to outlier rows. rCIL2-graph will not be very effective in this case. Notice that in this experiment, $r$ percentage of the images in each dataset is selected to be occluded with a size of $8\times 8$ block, and thus the decreasing curves of the clustering accuracy and NMI are flatter than those in Figures \ref{fig_yale_pixel} and \ref{fig_orl_pixel}.

The second experiment is conducted on a subset of AR dataset. This subset consists of 1,400 images from 100 subjects, 50 males and 50 females. These images are of non-occluded frontal views with various facial expressions in Sessions 1 and 2. For each image, we randomly select a local region to be replaced by an unrelated monkey image. The size of monkey image is $14\times 14$, i.e. about 20$\%$ pixels of each image are occluded. Figure \ref{fig_faceimages} (c) shows some face images with such unrelated image occlusions.

\begin{figure}[!t]
\centering
\includegraphics[width=0.40\textwidth]{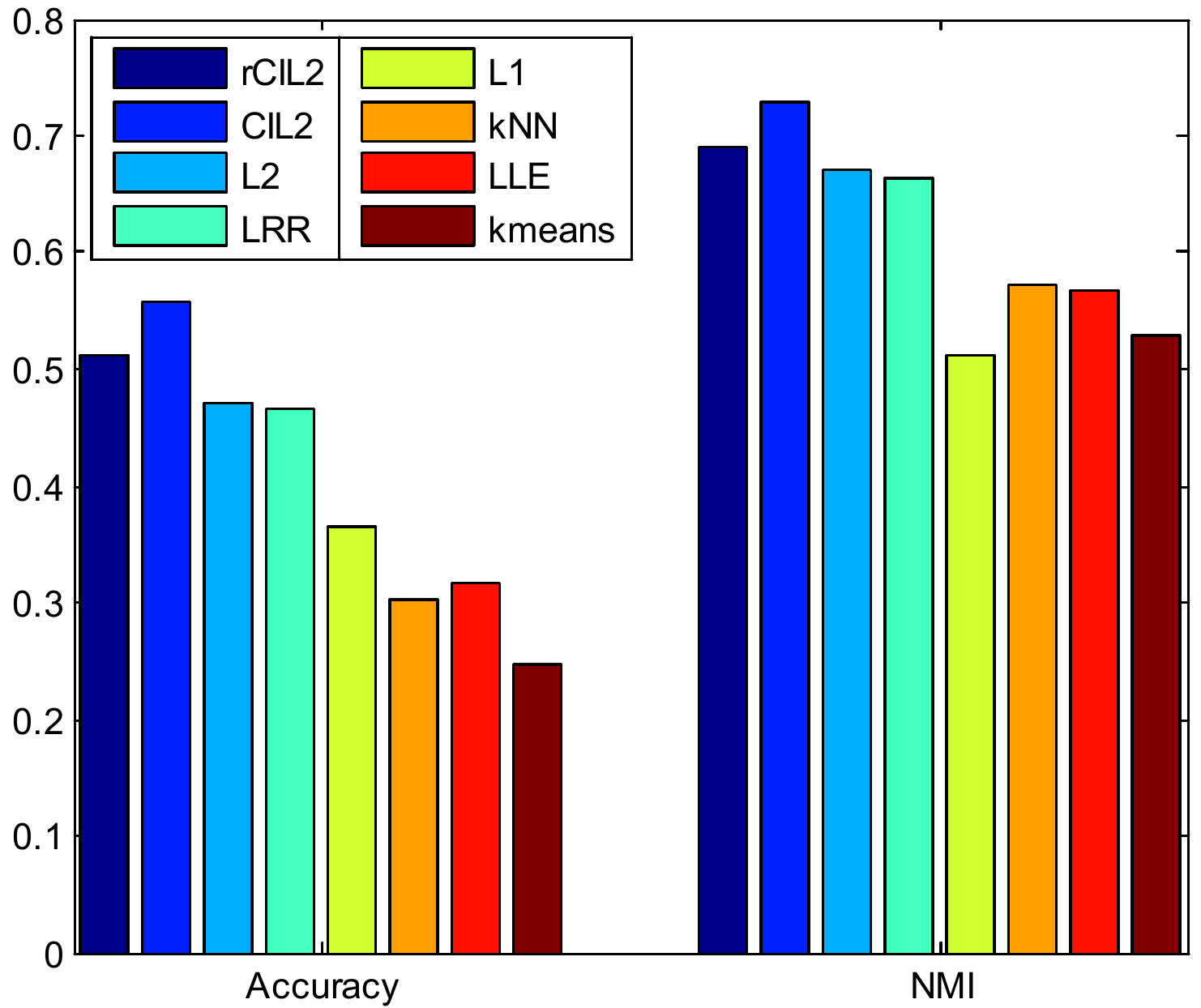}
\caption{Clustering accuracy and NMI on the AR dataset with an unrelated image occlusion for different algorithms.}
\label{fig_AR_imagecorr}
\end{figure}

Figure \ref{fig_AR_imagecorr} shows the clustering accuracy and NMI of each method on the AR dataset with unrelated monkey image occlusion. The experimental results are similar to the above experiment. Still, CIL2 obtains the best results, and rCIL2, LRR and L2-graph are competitive on this experiment.

\begin{table*}[!t]
\footnotesize
\centering
\caption{The clustering accuracy (\%) and NMI (\%)  of different algorithms on the AR dataset.}
\label{Tab_AR_sunscarf}
\centering
\begin{tabular}{c c c c c| c c c c }
\hline
\multirow{3}*{ Methods } & \multicolumn{4}{c}{Accuracy} & \multicolumn{4}{|c}{NMI} \\ \cline{2-9}
						 & \multicolumn{2}{c}{Session 1} & \multicolumn{2}{c}{Session 2} & \multicolumn{2}{|c}{Session 1} & \multicolumn{2}{c}{Session 2} \\ \cline{2-9}
						 & Sunglasses & Scarf & Sunglasses & Scarf & Sunglasses & Scarf & Sunglasses & Scarf\\ \hline
rCIL2	&\textbf{85.2}&	\textbf{78.4}	&\textbf{86.4}&	\textbf{81.2}&	\textbf{93.8}	&\textbf{90.1}&	\textbf{94.0}&	\textbf{91.5}\\
CIL2	&81.2&	75.4	&85.4	&79.0	&89.9	&87.9&93.8	&88.8\\
L2	&78.2&	71.6	&80.0	&72.6	&86.3&	83.8	&90.7&	83.8\\
LRR	&77.2&	72.2&	79.6&	74.6	&86.6&	84.7	&90.7&	84.7\\
L1	&43.8&	40.6& 27.8&	40.2	&72.3	&67.7	&55.8	&60.8\\
kNN	&26.4	&25.6	&26.8&	27.2&	65.4&	66.0&	66.1	&65.9\\
LLE	&28.0	&27.6&	33.2&	27.2&	63.4&	62.9	&66.6	&61.7\\
kmeans&	30.0&	29.4	&30.8	&29.8&	65.4&	63.8	&65.3	&65.5\\\hline
\end{tabular}
\end{table*}

\subsection{Results on Real-World Malicious Occlusion}

In real-world face recognition systems, people may wear sunglasses or scarfs which make the classification or clustering more challenging. In this subsection, we evaluate the robustness of the proposed method on the AR dataset with sunglasses and scarf occlusions. The AR dataset contains two separate sessions. In each session, each subject has 7 face images with different facial variations, 3 face images with sunglasses occlusion and 3 face images with scarf occlusion. Figure \ref{fig_faceimages} (d) shows some face images with such an occlusion. In each session, we conduct two experiments corresponding to the sunglasses and scarf occlusions. For sunglasses occlusion, we use the first 2 normal face images and 3 face images with sunglasses of each subject. For scarf occlusion, we use the first 2 normal face images and 3 face images with scarf of each subject.

Table \ref{Tab_AR_sunscarf} shows the clustering results on the AR dataset for the images with sunglasses and scarf occlusions. Different from the above experiments, rCIL2 achieves the best clustering accuracy and NMI in all cases. That is because the face images with sunglasses and scarf occlusions contain many outlier rows/features, and rCIL2 is designed for such a task. Both LRR and L2 graphs perform better than L1 graph, which is consistent with the result in \cite{LRRpami,LSR}.

\section{Conclusions}
In this paper, we study the robust subspace clustering problem, and present a general framework from the viewpoint of half-quadratic optimization to unify the L1 norm, Frobenius norm, L21 norm and nuclear norm based subspace clustering methods. Previous iteratively reweighted least squares optimization methods for the sparse and low rank minimization can be regarded as the half-quadratic optimization. As a new special case, we use the correntropy as the loss function for robust subspace clustering to handle the non-Gaussian and impulsive noises. An alternate minimization algorithm is used to optimize the non-convex correntropy objective. Extensive experiments on the face clustering with various types of corruptions and occlusions well demonstrate the effectiveness and robustness of the proposed methods by comparing with the state-of-the-art subspace clustering methods.

\section*{Acknowledgements}
This research is supported by the Singapore National Research Foundation under its International Research Centre @Singapore Funding Initiative and administered by the IDM Programme Office. Z. Lin is supported by National Natural Science Foundation of China (Grant nos. 61272341, 61231002, and 61121002).

{\footnotesize
\bibliographystyle{ieee}
\bibliography{CIL2}
}


\end{document}